%% file: main.tex
\documentclass{article}

\usepackage{arxiv}

\usepackage[utf8]{inputenc} 
\usepackage[T1]{fontenc}    
\usepackage{hyperref}       
\usepackage{url}            
\usepackage{amsfonts}       
\usepackage{nicefrac}       
\usepackage{microtype}      
\usepackage{graphicx}
\usepackage[table]{xcolor}

\usepackage{natbib}
\usepackage{algorithm, algpseudocode}
\usepackage{enumitem}

\usepackage{bm}
\usepackage{amsmath, amsthm, mathtools}

\usepackage{caption}
\usepackage{array}
\usepackage{adjustbox}
\usepackage{booktabs}
\usepackage{subcaption}
\usepackage[skip=10pt]{caption}

\usepackage{authblk}

\title{\LARGE PARIC: Probabilistic Attention Regularization \\for Language Guided Image Classification \\from Pre-trained Vison Language Models}

\author[1]{Mayank Nautiyal}
\author[2]{Stela Arranz Gheorghe}
\author[2]{Kristiana Stefa}
\author[1]{Li Ju}
\author[1]{Ida-Maria Sintorn}
\author[1,3]{Prashant Singh}

\affil[1]{Department of Information Technology, Uppsala University, Uppsala, Sweden}
\affil[2]{IT University of Copenhagen, Copenhagen, Denmark}
\affil[3]{Science for Life Laboratory, Uppsala University, Uppsala, Sweden}

\begin{document}
\maketitle

\input{sec/0_abstract}    
\input{sec/1_intro}

\input{sec/2_background}

\input{sec/4_methodology}

\input{sec/5_experiments}

\input{sec/6_discussion}
\input{sec/7_conclusion}

\section*{Acknowledgments}
The computations/data handling were enabled by the Berzelius resource provided by the Knut and Alice Wallenberg Foundation at the National Supercomputer Centre and by the National Academic Infrastructure for Supercomputing in Sweden (NAISS) at Chalmers e-Commons at Chalmers. P.S. acknowledges support from the Swedish Research Council through grant no. 2023-05593, and the Knut and Alice Wallenberg foundation through the Program for Academic leaders in Life Science (PALS). L.J. acknowledges funding from the Centre for Interdisciplinary Mathematics at Uppsala University and NAISS through Project 2024/22-1358.

\bibliographystyle{plainnat}

\bibliography{references}

\end{document}

%% file: sec/0_abstract.tex
\begin{abstract}
Language-guided attention frameworks have significantly enhanced both interpretability and performance in image classification; however, the reliance on deterministic embeddings from pre-trained vision-language foundation models to generate reference attention maps frequently overlooks the intrinsic multivaluedness and ill-posed characteristics of cross-modal mappings. To address these limitations, we introduce PARIC, a probabilistic framework for guiding visual attention via language specifications. Our approach enables pre-trained vision-language models to generate probabilistic reference attention maps, which align textual and visual modalities more effectively while incorporating uncertainty estimates, as compared to their deterministic counterparts. Experiments on benchmark test problems demonstrate that PARIC enhances prediction accuracy, mitigates bias, ensures consistent predictions, and improves robustness across various datasets.
\end{abstract}

%% file: sec/1_intro.tex
\section{Introduction}
\label{sec:intro}

Developing robust image classification models that generalize effectively to unseen or out-of-distribution data remains a challenging problem in computer vision. This issue largely arises from biases and limited diversity in training datasets \cite{5995347}. Standard models trained on such data often prioritize irrelevant background or contextual cues over the discriminative visual features that define each class \cite{Ribeiro16}. Consequently, these models struggle to generalize to unfamiliar or atypical examples, undermining their reliability and practical utility in real-world applications.

Learning robust joint representations for vision and language is an important challenge in modern deep learning research, where the goal is to construct a function $ f(\mathbf{V}, \mathbf{L}) $ that aligns visual data $ \mathbf{V} $ and linguistic data $ \mathbf{L} $ into a unified representation capturing shared semantics while preserving modality-specific details; mathematically, this can be expressed as $ f : \mathcal{V} \times \mathcal{L} \to \mathcal{Z} $, where $ \mathcal{Z} $ denotes the joint latent space encoding these semantics, with the primary challenge being to construct $ f $ such that it is both expressive and generalizable across diverse input types.
While humans seamlessly combine information across modalities to make complex inferences (e.g., understanding product reviews by interpreting customer-uploaded images alongside customer written feedback describing product experiences), traditional machine learning approaches often process each modality in isolation, making it challenging to exploit the rich complementary information inherent across modalities. Vision-Language Models (VLMs)  \cite{Radford21,Jia2021ScalingUV, li2022blip} address this limitation by training a unified model on large-scale image-text pairs, enabling the formation of rich cross-modal semantic representations. These learned associations subsequently drive performance across diverse downstream applications, including image captioning, visual question answering, content moderation, context-sensitive retrieval, and multimodal sentiment analysis \cite{Du2022ASO}.

Although pretrained VLMs excel at zero-shot classification—enabling inference on new data without additional training, a promising research direction lies in leveraging these large-scale pretrained models to guide smaller task-specific classifiers tailored to particular applications. A seminal effort in this area is the GALS framework \cite{GALS}, which proposes using language-based guidance from VLMs to steer task-specific classifiers through the generation of reference or guiding attention maps. Specifically, GALS transforms conventional classification labels into descriptive textual prompts, constructing semantically rich image-text pairs compatible with the VLM’s multimodal processing pipeline. These pairs are used to extract class-focused image attention maps, which highlight image regions semantically aligned with the corresponding linguistic descriptions. With these VLM-generated attention maps, a regularization mechanism is designed for the task-specific classifier, constraining its focus to align with semantically meaningful regions identified through language-guided reasoning. This approach enhances the interpretability, robustness, and accuracy of image classifiers by clarifying visual-textual relationships and reducing reliance on irrelevant visual cues.

While GALS improves classification accuracy by integrating pretrained VLMs as a guidance mechanism, it inherits a fundamental limitation shared by most conventional VLMs: their reliance on deterministic embeddings. Such embeddings struggle to account for the inherent ambiguities and uncertainties in multimodal data \cite{Blundell15, Oh2018ModelingUW}. Specifically, this  framework overlooks the fact that a single textual description can correspond to multiple distinct images, just as a single image may be accurately described by various textual expressions. This constraint hinders model effectiveness, particularly in tasks requiring fine-grained distinctions or generalization across diverse data distributions. To address these challenges, probabilistic VLMs have been developed  \cite{ProbVLM,PostHocVLM}, focusing on uncertainty estimation. Probabilistic embeddings provide a more nuanced representation by representing multimodal data as probability distributions rather than fixed points in the embedding space. Explicitly modelling uncertainty enables these frameworks to capture the variability in visual-textual relationships, leading to significant advancements in fine-grained classification, active learning, domain adaptation, and targeted model refinement via uncertainty quantification.

To accommodate multivaluedness and the ill-posed nature inherent to cross modal mappings, we introduce \textbf{PARIC}: \textit{``a probabilistic framework for guiding visual attention using language specifications"}. PARIC leverages probabilistic VLMs to generate uncertainty-aware attention maps from descriptive textual prompts derived from classification labels. These probabilistic attention maps explicitly guide a separate task-specific neural classifier to focus on semantically meaningful visual regions, aiding model interpretability, robustness, and generalization. Building upon GALS \cite{GALS} and probabilistic adapters \cite{ProbVLM}, our approach bridges the gap between deterministic and probabilistic multimodal paradigms. We demonstrate improvements in accuracy, robustness, and interpretability across benchmark visual classification datasets, underscoring the efficacy of uncertainty-aware multimodal guidance.

%% file: sec/2_background.tex
\section{Background}
\label{sec:background}

Vision-Language Models (VLMs) have been pivotal in multimodal research, enabling simultaneous processing of images and text within a unified architecture. Foundational models like CLIP \cite{Radford21} and ALIGN \cite{Jia2021ScalingUV} employ contrastive learning on large-scale datasets (e.g., LAION-400M \cite{Schuhmann21}), to align image-text pairs in a shared embedding space, excelling in tasks such as cross-modal retrieval and zero-shot classification \cite{Wang2016ACS}. Unified pre-training approaches such as BLIP \cite{li2022blip}, combine contrastive and generative objectives, enhancing performance in tasks like image captioning and visual question answering. However, a notable limitation of these models lies in their reliance on deterministic embeddings. Although suitable for many downstream applications, deterministic representations may not adequately address the uncertainties or variations inherent in real-world data, especially when dealing with fine-grained distinctions, unseen categories, or tasks that demand high precision and adaptability \cite{Yang2021ProbabilisticMO}.

\paragraph{Probabilistic Embeddings} have recently been explored in various workflows \cite{ProbVLM, Chun2021ProbabilisticEF, 10.5555/3600270.3601137, Neculai2022ProbabilisticCE}, and hold the potential to address the inherent ambiguity and multivaluedness in VLMs by explicitly modeling uncertainty, offering a more nuanced representation of complex multimodal data \cite{Blundell15, Oh2018ModelingUW}. However, training probabilistic VLMs from scratch demands substantial computational resources and large scale datasets \cite{Stirn22}. Recent advancements address this limitation through post-hoc probabilistic methods. Specifically, Bayesian posterior approximation, as seen in \cite{PostHocVLM}, estimates uncertainty over the final VLM layers without retraining. Alternatively, \cite{ProbVLM} introduces lightweight probabilistic adapters to convert deterministic embeddings into probabilistic embeddings, offering a computationally efficient approach that leverages pre-trained models.

\paragraph{Language as a Guide for Visual Models} based on attention mechanisms allow models to focus selectively on the most salient parts of an input, such as specific regions in an image \cite{GAN2024122731}. In multimodal settings, language-guided attention facilitates the alignment of visual regions with semantic textual cues, enabling the generation of more robust and interpretable attention maps. Using language to direct attention, models can prioritize critical visual features while suppressing irrelevant background information, thereby reducing the risk of learning spurious correlations. For example, this approach ensures that a model associates a bird's species with its defining characteristics rather than unrelated contextual factors like its habitat. Recent frameworks, such as GALS \cite{GALS}, have demonstrated the efficacy of modulating visual attention using language cues—a principle that also forms the foundation of this work.

\paragraph{Information Grounding in VLMs} refers to the process of associating textual descriptions with specific regions or attributes of visual inputs \cite{inforgrounding1, inforgrounding2}, enabling models to better understand the relationship between text and fine-grained visual features for improved accuracy and contextual awareness in predictions. This is typically achieved using pre-trained VLMs like CLIP or BLIP, which align visual and textual data in a shared embedding space to inform downstream tasks such as classification, segmentation, or retrieval. However, grounding becomes challenging in scenarios with mismatches between linguistic and visual modalities \cite{infoground3}, particularly in fine-grained tasks where subtle distinctions are crucial, highlighting the need for robust mechanisms to ensure semantically meaningful correspondences between text and image.

\paragraph{Supervising Attention in Visual Models} Attention mechanisms plays an important role in identifying the most relevant regions of an image for a specific task that the model should focus on during training \cite{Selvaraju19}. By explicitly supervising attention, models can be guided to prioritize the most relevant input areas, leading to enhanced performance on tasks requiring fine-grained discrimination. Approaches such as the ``Right for the Right Reasons" \cite{Ross17} enforce alignment between attention and task-specific requirements, ensuring the model focuses on appropriate visual features rather than spurious correlations. This supervision can be further strengthened by leveraging saliency maps derived from pre-trained vision-language models, which highlight semantically meaningful regions in images. In frameworks like GALS, attention supervision is integrated with language-guided cues to create robust models capable of handling both high-level task specifications and detailed visual features, improving interpretability and task accuracy.

%% file: sec/4_methodology.tex
\section{Problem Formulation}
\label{sec:method}

\begin{figure*}[t]
  \centering
  \includegraphics[width=0.8\textwidth]{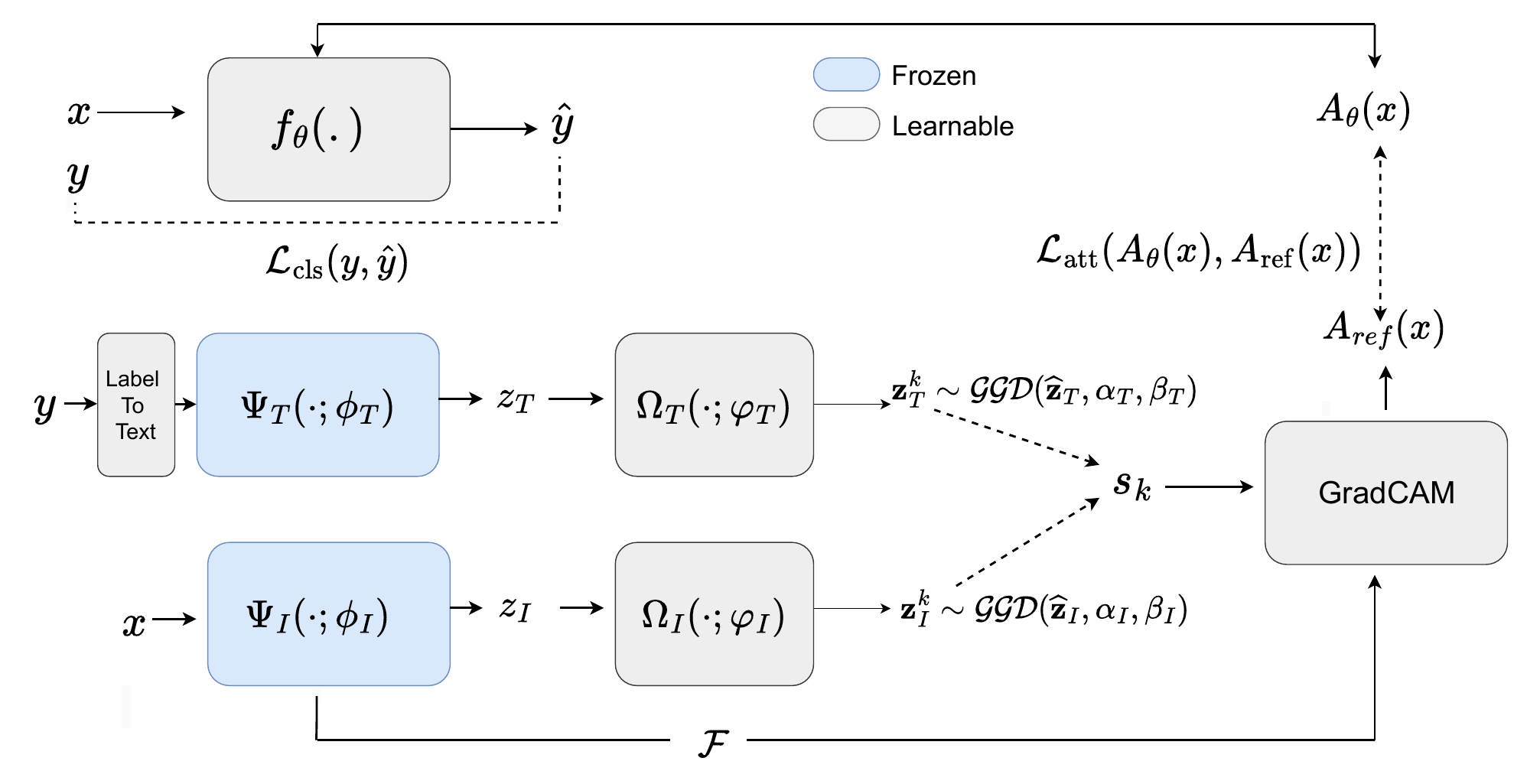}
\caption{\textbf{\textit{PARIC Workflow.}}  
The classifier $f_{\theta}$ predicts the label $\hat{y}$ and generates an attention map $A_{\theta}(x)$ for input image $x$. The ProbVLM pipeline expects image-text pairs, so the label $y$ is first converted into text prompts. These text prompts, along with the image, $x$, are processed by frozen CLIP encoders $\Psi_T$ (text) and $\Psi_I$ (image) to produce deterministic embeddings $\mathbf{z}_T$ and $\mathbf{z}_I$. Trainable adapters $\Omega_T$ and $\Omega_I$ model these embeddings as Generalized Gaussian Distributions (GGDs), from which $K$ samples are drawn to compute similarity scores. Grad-CAM combines these scores with CLIP's image feature map $\mathcal{F}$ to generate $K$ saliency maps, which are aggregated using mean or median into a reference map $A_{\mathrm{ref}}(x)$. This map guides $A_{\theta}(x)$ via $\mathcal{L}_{\mathrm{att}}$, complementing $\mathcal{L}_{\mathrm{cls}}$ to improve the robustness and interpretability of the classifier.}

\label{pipeline}
\end{figure*}

Let \(\mathcal{X} \in \mathbb{R}^{H \times W \times C}\) be the input space of images, where \(H\) and \(W\) denote spatial dimensions and \(C\) the number of channels, and let \(\mathcal{Y}\) be be the set of class labels. Consider a neural network classifier \(f_{\theta} : \mathcal{X} \to \mathcal{Y}\), parameterized by \(\theta\), which generates spatial attention map \(A_{\theta}(x) \in \mathbb{R}^{H \times W}\) for each image \(x \in \mathcal{X}\). Our goal is to regularize \(A_{\theta}(x)\) by leveraging a probabilistic attention prior derived from a frozen CLIP encoder, realized through probabilistic adapters akin to ProbVLM.

\paragraph{Probabilistic Encoder Integration}
Pre-trained zero-shot models like CLIP are adapted to generate probabilistic embeddings, modeling image-text alignments as random variables rather than fixed points. With probabilistic adapters, any image or caption \(\zeta\) is mapped to a random variable \(\mathbf{z}\) following a Generalized Gaussian Distribution (GGD), parameterized by the output of the probabilistic adapter \(\Omega(\cdot; \varphi)\) applied to the frozen CLIP encoder \(\Psi(\cdot; \phi)\), where \(\varphi\) and \(\phi\) denote the parameters of the adapter and the CLIP encoder, respectively. Formally, for any image or caption \(\zeta\), its embedding \(\mathbf{z}\), the corresponding probability density function (PDF), and the parameters of the distribution are given by:
\begin{align}
\mathbf{z} &\sim \mathcal{GGD}(\widehat{\mathbf{z}},\alpha,\beta),\\
p(\mathbf{z};\widehat{\mathbf{z}},\alpha,\beta) &\propto \exp\bigl(-|\mathbf{z}-\widehat{\mathbf{z}}|^\beta / \alpha^\beta\bigr),\\
(\widehat{\mathbf{z}},\alpha,\beta) 
&= \left[ \Omega(\cdot; \varphi) \circ \Psi(\cdot; \phi)\right](\zeta),
\end{align}
where $\mathcal{GGD}(\cdot)$ denotes a generalized Gaussian distribution, $\widehat{\mathbf{z}} \in \mathbb{R}^d$ is the mean embedding with dimension $d$ and $\alpha,\beta\in\mathbb{R}^{+}$ capture scale and shape of the embedding distribution. This design reflects both aleatoric (via $\alpha,\beta$) and epistemic (via Monte Carlo dropout) uncertainties. The adapters use intra-modal and cross-modal alignment objectives to ensure consistent image-text embeddings.

\paragraph{Sampling from Probabilistic Embeddings} To account for the uncertainty inherent in the multimodal task, we sample \(K\) points from the learned image and text probabilistic distributions, i.e., \(\mathbf{z}_I^{(k)} \sim \mathcal{GGD}(\widehat{\mathbf{z}}_I,\alpha_I,\beta_I)\) and \(\mathbf{z}_T^{(k)} \sim \mathcal{GGD}(\widehat{\mathbf{z}}_T,\alpha_T,\beta_T)\). Each sample represents a different instantiation of the embeddings, reflecting distinct possible interpretations of the image-text pair.

\paragraph{Saliency Map Generation and Aggregation} To localize semantically relevant image regions, we generate saliency maps from $K$ sampled probabilistic embeddings. For each sample $k$, we compute a similarity score $s^{(k)} = \mathbf{z}_I^{(k)} \cdot \mathbf{z}_T^{(k)}$ between the image and text embeddings. Using this score and the image encoder's final convolutional feature map $\mathbf{\mathcal{F}} \in \mathbb{R}^{H \times W \times C}$, we derive attention maps via Grad-CAM~\cite{8237336}. Channel-wise weights $\{w_c^{(k)}\}_{c=1}^C$ are obtained by spatial averaging of the gradients $\partial s^{(k)} / \partial \mathbf{\mathcal{F}}$. The saliency map $A^{(k)}(x)$ is then constructed as a weighted sum of feature map channels, with ReLU activation ensuring non-negative saliency values: \(
A^{(k)}(x) = \mathrm{ReLU} \left( \sum_{c=1}^C w_c^{(k)} \mathbf{\mathcal{F}}_c \right)
\). This results in a set of $ K $ saliency maps $\{A^{(k)}(x)\}_{k=1}^K$, which are consolidated into a robust reference attention map $ A_{\mathrm{ref}}(x)$. To achieve this consolidation, we explore two distinct aggregation strategies: mean aggregation, which produces a smooth central tendency, and median aggregation, which provides an outlier-resistant summary of the sampled maps. Consequently, we define two PARIC model variants:
\begin{equation}
A_{\mathrm{ref}}(x) = \begin{cases} \frac{1}{K} \sum_{k=1}^K A^{(k)}(x) & \text{(PARIC mean)} \\ \mathrm{median}(\{A^{(k)}(x)\}_{k=1}^K) & \text{(PARIC median)} \end{cases}.
\label{eq:attention}
\end{equation}

\paragraph{Guiding the Visual Classifier} To promote alignment between the classifier's attention map $A_{\theta}(x)$ and the reference attention map $A_{\text{ref}}(x)$, we introduce an attention regularization term:
\begin{equation}
\label{eq:att-penalty}
\mathcal{L}_{\mathrm{att}}(A_{\theta}(x), A_{\mathrm{ref}}(x)) 
\;=\;
\sum_{i,j}\bigl|(1 - A_{\mathrm{ref}}(x)_{i,j}) \cdot A_{\theta}(x)_{i,j}\bigr|,
\end{equation}
where \((1 - A_{\mathrm{ref}}(x))\) represents regions deemed irrelevant by the reference attention map, thus discouraging the classifier from attending to these areas. We combine this with the standard cross-entropy classification loss:

\begin{equation}
\label{eq:cls}
\mathcal{L}_{\mathrm{cls}}(f_{\theta}(x),y) 
\;=\; 
-\sum_{c=1}^C \mathbf{1}\{c=y\}\,\log\,p_{\theta}(c\!\mid\!x),
\end{equation}
where $p_{\theta}(c\mid x)$ is the predicted probability for class $c$. The final objective function for PARIC is then formulated as a weighted sum of the classification and the attention regularization loss:

\begin{equation}
\label{eq:total-loss}
\mathcal{L}_{\mathrm{total}}(x,y) 
\;=\;
\mathcal{L}_{\mathrm{cls}}(f_{\theta}(x),y)
\;+\;
\lambda\,\mathcal{L}_{\mathrm{att}}(A_{\theta}(x),A_{\mathrm{ref}}(x)),
\end{equation}
where \(\lambda>0\) balances classification accuracy against attention alignment. Minimizing \(\mathcal{L}_{\mathrm{total}}\) guides the classifier towards stable, high-importance regions identified via the probabilistic embeddings, resulting in more interpretable and robust predictions. \autoref{pipeline} provides a schematic overview of the PARIC model.

%% file: sec/5_experiments.tex
\section{Experiments}
\label{sec:experiments}

We compare PARIC with the well-established GALS framework across three diverse datasets (MS-COCO \cite{cocodataset}, Waterbird \cite{Sagawa2019DistributionallyRN}, Food101 \cite{bossard14}) over five independent trials to evaluate whether integration of probabilistic encoders leads to improved classification accuracy and better generalization performance in task-specific classifiers. We assess two variants of PARIC— \textit{PARIC-mean} and \textit{PARIC-median} which differ in their aggregation strategies for combining attention maps derived from sampled embeddings. We outline the specific implementation details below.

\paragraph{Model Architecture}
We adopt CLIP with a ResNet50 backbone, pre-trained on ImageNet, as the foundation of our vision-language model. CLIP is selected for its excellent performance in aligning vision and language representations, as well as its architectural compatibility with the existing GALS framework \cite{GALS}. This compatibility ensures a consistent baseline, allowing us to isolate and evaluate the effects of incorporating probabilistic layers. For the downstream classification task, we use the same image classifier architecture as in the GALS framework \cite{GALS}. 

\paragraph{Probabilistic Adapters}
We augment CLIP's deterministic encoders with post-hoc probabilistic adapters, implemented as small MLPs with dropout layers following the structure outlined in \cite{ProbVLM}. These adapters transform embeddings into parameters of a Generalized Gaussian Distribution, enabling uncertainty modeling in image-text alignments.

\paragraph{Attention Map Generation}
To extract attention via language specification, we employ prompts similar to those used in CLIP (e.g., “an image of \textit{category}” or “a photo of \textit{category}”), where \textit{category} corresponds to task-relevant concepts. For each image-text pair, 50 embeddings are sampled from the learned probabilistic distribution. These embeddings are processed through GradCAM \cite{8237336}, applied at the final convolutional layer (Layer 4) of the ResNet-50 CLIP encoder, to produce individual saliency maps highlighting the most relevant image regions. The resulting saliency maps are aggregated using either mean or median pooling to create a representative attention map for each image-text pair with uncertainty estimates, which can be calculated by pixel-wise standard deviation of the map samples. a These aggregated maps are then used to regularize the attention mechanisms of task-specific classifiers with the aim of ensuring alignment with semantically meaningful regions.

\paragraph{Implict vs Explicit Dataset Bias}
We examine two variants of the Waterbirds dataset to study explicit bias. The first, Waterbirds-$100\%$, enforces perfect correlation between bird type and background during training (e.g., waterbirds on water, landbirds on land). The second, Waterbirds-$95\%$, introduces $5\%$ of training samples that break this correlation, testing robustness to slight bias deviations. For dataset details, see \cite{Sagawa2019DistributionallyRN}.

The Food-101 dataset is used to study implicit biases from its uncurated training set, which includes noisy labels, co-occurring elements, and visual artifacts. For instance, certain ingredients (e.g., sauces) are spuriously correlated with specific dishes, introducing biases. The evaluation set, however, is curated, creating a distribution shift. We use two setups: \textit{Red Meat Subset}: following \cite{GALS}, we use a five-way classification task (baby back ribs, filet mignon, pork chop, prime rib, steak) to predict red meat categories. \textit{Meat Subset}: We introduce a three-way classification task (red meat, white meat, fish) by filtering 50 animal-based meat classes, analyzing biases like co-occurring ingredients and image quality variations.

For MS-COCO \cite{Lin15}, we extract a subset of images labeled with the ``Person" category, we further restrict this subset to images included in the MS-COCO-ApparentGender dataset, which introduces implicit bias through gender labels inferred from captions (``Man," ``Woman," or ``Person"). This filtering reduces the size of the train, validation, and test sets, allowing us to study the impact of implicit bias on classification performance. Implicit bias in the MSCOCO-ApparentGender setup stems from the gender labels which are based on outward appearances described in image captions rather than objective features of individuals. This reliance on subjective descriptions introduces societal stereotypes into the dataset, where certain activities, clothing, or contexts are disproportionately associated with specific genders (e.g., sports with men or domestic settings with women). Additionally, MS-COCO itself is inherently biased, having an unbalanced $1:3$ women to men ratio \cite{Tang21}.  Such under-representation amplifies the risk that the model learns spurious correlations from majority-class features, favoring men over women in its predictions. This imbalance, combined with the subjective nature of the apparent gender labels and the underuse of neutral terms like “Person,” aggravates the model’s tendency to internalize and reinforce societal stereotypes, influencing its performance across gender categories.

\subsection{Results}

\paragraph{Explicit Bias on Waterbirds}
On the Waterbirds-$100\%$ dataset, \textit{PARIC-Median} achieves an overall accuracy of ($96.80\%$), followed by \textit{PARIC-Mean} ($96.69\%$) and GALS ($96.65\%$) (\autoref{tab:waterbirds}). GALS achieves the highest accuracy for the Waterbird class ($95.67\% \pm 0.71$), but also exhibits higher variance compared to both \textit{PARIC-Mean} ($\pm 0.33$) and \textit{PARIC-Median} ($\pm 0.66$). For the Landbird class, \textit{PARIC-Mean} achieves the best accuracy ($97.08\% \pm 0.50$), with lower variance ($57\%$ reduction) compared to GALS ($96.81\% \pm 1.16$). These results highlight \textit{PARIC}'s ability to reduce variance and improve stability, particularly in the presence of strong background-label correlations.

On the more challenging Waterbirds-$95\%$ dataset, where $5\%$ of training samples break the background-label correlation, GALS achieves the highest overall ($96.93\%$) accuracy. \textit{PARIC-Mean} achieves the best Waterbird accuracy ($94.66\% \pm 0.87$), while \textit{PARIC-Median} demonstrates the lowest variance for the Waterbird class ($\pm 0.32$) and overall ($ \pm 0.15$). \autoref{tab:waterbirds95} indicates that \textit{PARIC-Median} is more robust to outliers and provides more stable predictions, even when the dataset contains counterexamples to the dominant bias. Notably, both GALS and PARIC do not perform worse for the Waterbird class on Waterbird-100 which is more challenging, as compared to Waterbird-95. This could be attributed to a combination of regularization via noise, a greater diversity of training data samples in Waterbirds-100 or the noise affecting non-critical regions of the test images, which the attention mechanism is able to mitigate.

Overall, \textit{PARIC} demonstrates consistent performance across both datasets, with \textit{PARIC-Median} showing particular strength in reducing variance and improving robustness. GALS performed well on the Landbird class in the Waterbirds-$95\%$ dataset, and Waterbird class in the Waterbirds-$100\%$ dataset indicating gains due to more model expressiveness (and less regularization).

\begin{table}[h!]
\centering
\setlength{\tabcolsep}{5pt}
\renewcommand{\arraystretch}{1.25}

\begin{subtable}[t]{\columnwidth}
    \centering
    \subcaption{Waterbirds 100\%}
    \begin{adjustbox}{max width=\columnwidth}
    \begin{tabular}{c  c  c  c}

    \toprule
    \textbf{Method} & \textbf{Waterbird} & \textbf{Landbird} & \textbf{Overall} \\
    \midrule
    GALS         & \textbf{95.67} \(\pm\) {\footnotesize 0.7101} & 96.81 \(\pm\) {\footnotesize 1.1617} & 96.65 \(\pm\) {\footnotesize 0.9319} \\
    PARIC mean   & 95.04 \(\pm\) {\footnotesize \textbf{0.3292}} & \textbf{97.08} \(\pm\) {\footnotesize \textbf{0.5003}} & 96.69 \(\pm\) {\footnotesize 0.5195} \\
    PARIC median & 95.24 \(\pm\) {\footnotesize 0.6620} & 96.92 \(\pm\) {\footnotesize 0.6690} & \textbf{96.80} \(\pm\) {\footnotesize \textbf{0.5059}} \\
    \bottomrule
    \end{tabular}
    \end{adjustbox}
    \label{tab:waterbirds}
\end{subtable}

\vspace{1em} 

\begin{subtable}[t]{\columnwidth}
    \centering
    \subcaption{Waterbirds 95\%}
    \begin{adjustbox}{max width=\columnwidth}
    \begin{tabular}{c c c c}

    \toprule
    \textbf{Method} & \textbf{Waterbird} & \textbf{Landbird} & \textbf{Overall} \\
    \midrule
    GALS         & 93.69 \(\pm\) {\footnotesize 0.6691} & \textbf{97.46} \(\pm\) {\footnotesize 0.2487} & \textbf{96.93} \(\pm\) {\footnotesize 0.1870} \\
    PARIC mean   & \textbf{94.66} \(\pm\) {\footnotesize 0.8683} & 96.90 \(\pm\) {\footnotesize 0.5459} & 96.59 \(\pm\) {\footnotesize 0.3582} \\
    PARIC median & 94.13 \(\pm\) {\footnotesize \textbf{0.3220}} & 97.35 \(\pm\) {\footnotesize \textbf{0.2180}} & 96.91 \(\pm\) {\footnotesize \textbf{0.1513}} \\
    \bottomrule
    \end{tabular}
    \end{adjustbox}
    \label{tab:waterbirds95}
\end{subtable}

\caption{PARIC vs. GALS: Classification accuracy on (a) Waterbirds 100\%  and (b) Waterbirds 95\% test sets (mean $\pm$ std. dev.).}
\label{tab:waterbirds_merged}
\end{table}

\begin{table*}[htbp]
\centering
\setlength{\tabcolsep}{5pt}
\renewcommand{\arraystretch}{1.25}
\begin{adjustbox}{max width=\textwidth}
\begin{tabular}{c c c c c}

\toprule
\textbf{Method} & \textbf{Man} & \textbf{Woman} & \textbf{Overall} & \textbf{Outcome Divergence} \\
\midrule
GALS & 73.72 \(\pm\) {\footnotesize 8.9108} & 64.15 \(\pm\) {\footnotesize 8.0289} & 68.93 \(\pm\) {\footnotesize 2.0765} & 0.0729 \(\pm\) {\footnotesize 0.0406} \\
PARIC mean & 75 \(\pm\) {\footnotesize 5.6892} & \textbf{65.42} \(\pm\) {\footnotesize 5.5482} & \textbf{70.21} \(\pm\) {\footnotesize 1.8194} & \textbf{0.0597} \(\pm\) {\footnotesize 0.0249} \\
PARIC median & \textbf{75.21} \(\pm\) {\footnotesize 4.7765} & 63.93 \(\pm\) {\footnotesize \textbf{2.7205}} & 69.57 \(\pm\) {\footnotesize 2.8968} & 0.0642 \(\pm\) {\footnotesize \textbf{0.0119}} \\
\bottomrule
\end{tabular}
\end{adjustbox}
\caption{\textbf{MSCOCO.} Classification accuracy of PARIC and the baseline GALS on the MS-COCO test set. Highlighted in \textbf{bold} are the best results in each column (mean $\pm$ std. dev.).}
\label{tab:coco}
\end{table*}

\textbf{Implicit bias on COCO Gender.} 
On the COCO Gender dataset, PARIC outperforms the baseline GALS in both overall accuracy and fairness (Table \autoref{tab:coco}). \textit{PARIC-Mean} achieves the highest overall accuracy ($70.21\%$), surpassing GALS ($68.93\%$), and improves classification accuracy for the underrepresented “Woman” class to $65.42\%$. PARIC also demonstrates greater stability and consistency, with \textit{PARIC-Mean} achieving the lowest standard deviation for overall accuracy ($\pm1.82$) and \textit{PARIC-Median} minimizing outcome divergence (Jensen-Shannon divergence \cite{lin1991divergence} calculated over the score distributions of the two classes; lower values indicate better performance) variability to $0.0119$, compared to GALS ($0.0406$). For the “Woman” class, \textit{PARIC-Median} further reduces variability to $(\pm2.72)$, ensuring more predictable and trustworthy outcomes. Additionally, PARIC reduces gender disparities, with \textit{PARIC-Mean} achieving a lower outcome divergence ($0.0597$) compared to GALS ($0.0729$), indicating more balanced performance across gender categories. 

\paragraph{Robustness to noisy data}
To evaluate PARIC's robustness, we assess its performance on the Food-101 dataset containing implicit biases and image noise. We test the model on two setups: (i) the Red Meat Subset with five classes and (ii) the Meat Dataset with three classes. These tasks assess the model's ability to handle challenging noisy scenarios while accounting for biases.

The experimental results for the Red Meat Subset (\autoref{tab:redmeat}) demonstrate that GALS achieves higher accuracy for 3 classes (Filet Mignon, Pork Chop, Prime Rib), while PARIC-mean achieves lower variance compared to PARIC-median and GALS. 
In the Steak category, GALS achieves an accuracy of $50.4\%$, which PARIC-mean improves this to $55.6\%$ while significantly reducing variability ($\pm 2.0861$). Across all categories, PARIC-mean exhibits lower variance in accuracy scores, enhanced model stability.

On the Meat Dataset (\autoref{tab:meat}), PARIC demonstrates its ability to handle class imbalance effectively. With class distributions of $38\%$ Fish, $36\%$ Red Meat, and $26\%$ White Meat, PARIC-median reduces prediction variability for the minority class (White Meat) from $0.5824$ (GALS) to $0.2486$. While PARIC shows slightly higher standard deviations for majority classes (Red Meat and Fish), it maintains higher overall accuracy ($85.90\%$ vs. $85.78\%$ for GALS). This suggests that PARIC explores a broader range of feature representations.

\begin{table*}[htbp]
    \centering
    \setlength{\tabcolsep}{5pt}
    \renewcommand{\arraystretch}{1.25}
    
    \begin{subtable}[t]{\textwidth}
        \centering
        \subcaption{Red Meat Dataset}
        \begin{adjustbox}{max width=\textwidth}
            \begin{tabular}{c c c c c c c}

                \toprule
                \textbf{Method} & \textbf{Baby Back Ribs} & \textbf{Filet Mignon} & \textbf{Pork Chop} & \textbf{Prime Rib} & \textbf{Steak} & \textbf{Overall} \\
                \midrule
                GALS         & 86.56 \(\pm\) {\footnotesize 1.1482} & \textbf{72.4} \(\pm\) {\footnotesize 1.9919} & \textbf{71.2} \(\pm\) {\footnotesize 1.9758} & \textbf{85.92} \(\pm\) {\footnotesize 1.1142} & 50.4 \(\pm\) {\footnotesize 3.4409} & 73.30 \(\pm\) {\footnotesize \textbf{0.4422}} \\
                PARIC mean   & \textbf{87.84} \(\pm\) {\footnotesize \textbf{0.8616}} & 69.52 \(\pm\) {\footnotesize \textbf{0.7756}}  & 70.2 \(\pm\) {\footnotesize \textbf{0.9666}}  & 84.32 \(\pm\) {\footnotesize \textbf{0.6881}}  & 55.6 \(\pm\) {\footnotesize \textbf{2.0861}}  & \textbf{73.50} \(\pm\) {\footnotesize 0.4837} \\
                PARIC median & 86.32 \(\pm\) {\footnotesize 1.5676}  & 70.96 \(\pm\) {\footnotesize 2.2712}  & 69.68 \(\pm\) {\footnotesize 1.0552}  & 84.40 \(\pm\) {\footnotesize 0.9797}  & \textbf{55.68} \(\pm\) {\footnotesize 2.4709}  & 73.41 \(\pm\) {\footnotesize 0.8654} \\
                \bottomrule
            \end{tabular}
        \end{adjustbox}
        \label{tab:redmeat}
    \end{subtable}
    
    \vspace{1em} 
    
    \begin{subtable}[t]{\textwidth}
        \centering
        \subcaption{Meat Dataset}
        \begin{adjustbox}{max width=\textwidth}
            \begin{tabular}{c c c c c}
                \toprule
                \textbf{Method} & \textbf{White meat} & \textbf{Red meat} & \textbf{Fish} & \textbf{Overall} \\
                \midrule
                GALS         & 85.37 \(\pm\) {\footnotesize 0.5824}  & 81.48 \(\pm\) {\footnotesize \textbf{0.2889}} & 89.12 \(\pm\) {\footnotesize \textbf{0.3978}} & 85.78 \(\pm\) {\footnotesize 0.2504} \\
                PARIC mean   & \textbf{85.72} \(\pm\) {\footnotesize 0.2608}  & 81.15 \(\pm\) {\footnotesize 0.5898}  & \textbf{89.33} \(\pm\) {\footnotesize 0.4999}  & \textbf{85.90} \(\pm\) {\footnotesize 0.2352} \\
                PARIC median & 84.84 \(\pm\) {\footnotesize \textbf{0.2486}}  & \textbf{81.49} \(\pm\) {\footnotesize 0.6732}  & 89.15 \(\pm\) {\footnotesize 0.8396}  & 85.61 \(\pm\) {\footnotesize \textbf{0.1469}} \\
                \bottomrule
            \end{tabular}
        \end{adjustbox}
        \label{tab:meat}
    \end{subtable}

    \caption{Comparison of classification accuracy of PARIC and the baseline GALS on (a) Read Meat Dataset; (b) Meat Dataset. Highlighted in \textbf{bold} are the best results in each column (mean $\pm$ std. dev.).}
    \label{tab:overall}
\end{table*}

\begin{figure}[htbp]
    \centering
    \includegraphics[width=0.48\columnwidth]{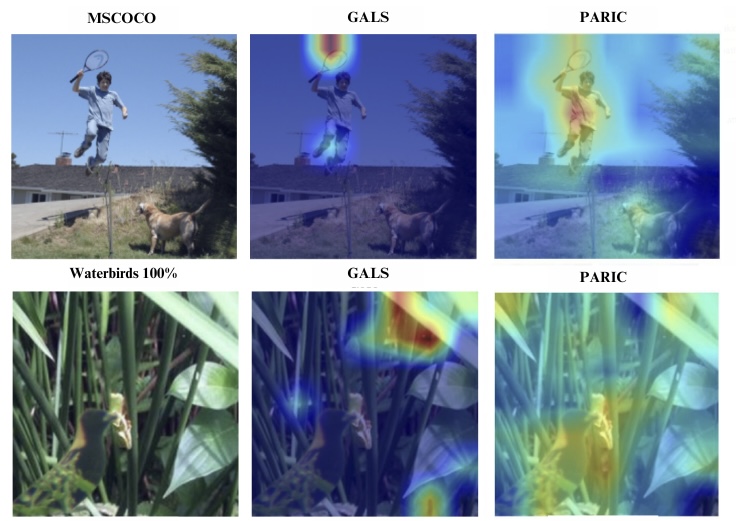}
    \includegraphics[width=0.48\columnwidth]{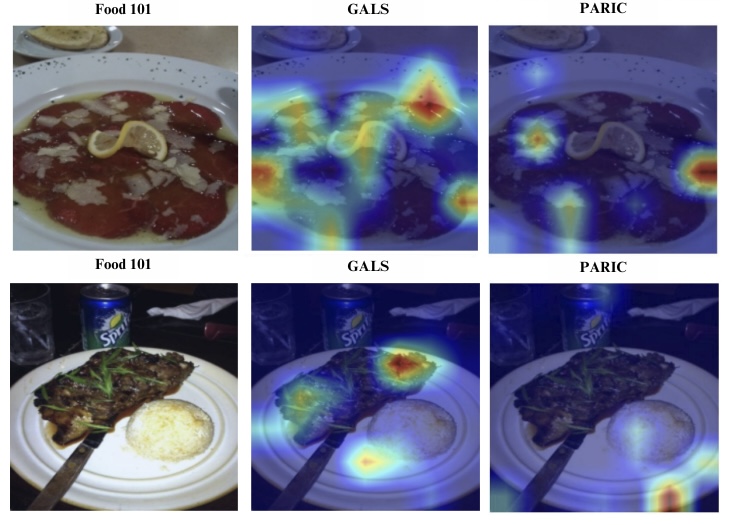}

    \caption{\textbf{\textit{Attention Map Visualization.}} for three instances: COCO and Waterbirds 100\% (GALS vs. PARIC Mean) and Food-101 (GALS vs. PARIC Median). Each row shows the original image, the attention map from frozen CLIP, and the refined map after integrating probabilistic layers. The first two instances show the strengths of the probabilistic approach, where the attention maps are more accurate, while the case of Food-101 shows an experiment where PARIC performs worse, with the regularization being too strong and limiting.}
    \label{attmaps}
\end{figure}


\begin{figure*}[ht]
    \centering
    \begin{subfigure}[t]{0.475\textwidth}
        \centering
        \includegraphics[width=\linewidth]{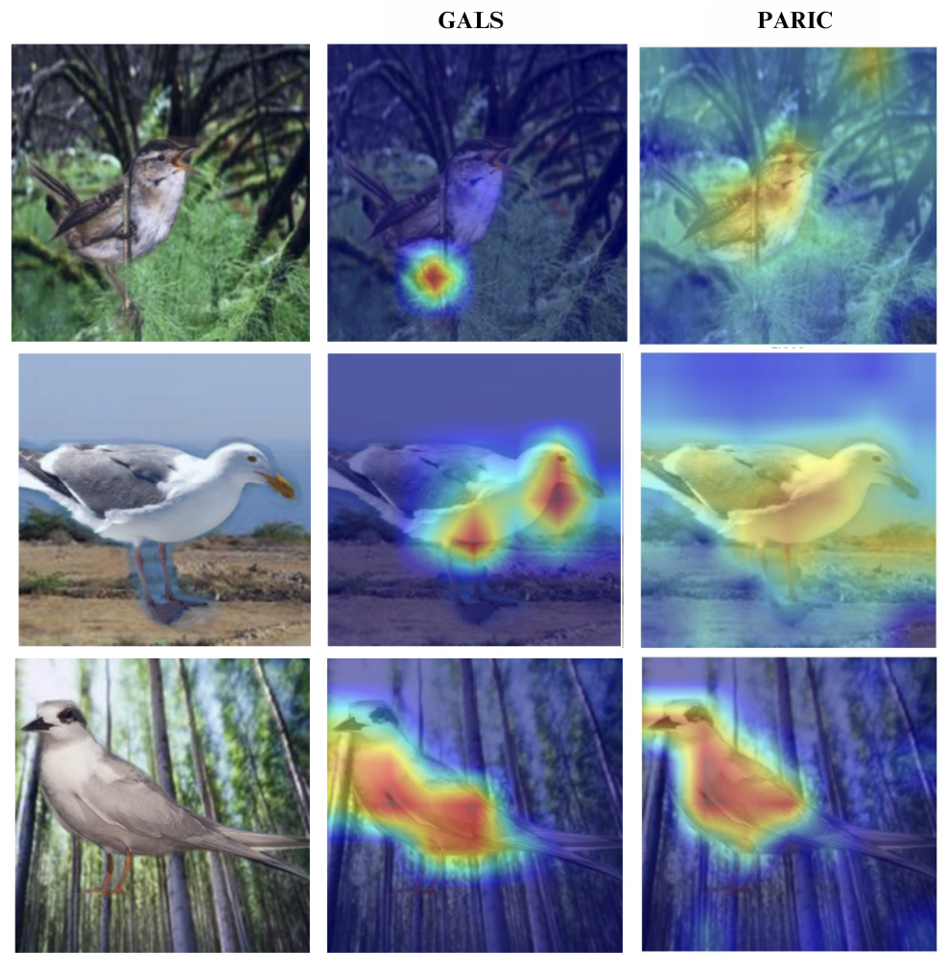}
        \subcaption{Effective attention maps.}
        \label{attmaps2}
    \end{subfigure}
    \hfill
    \begin{subfigure}[t]{0.5\textwidth}
        \centering
        \includegraphics[width=\linewidth]{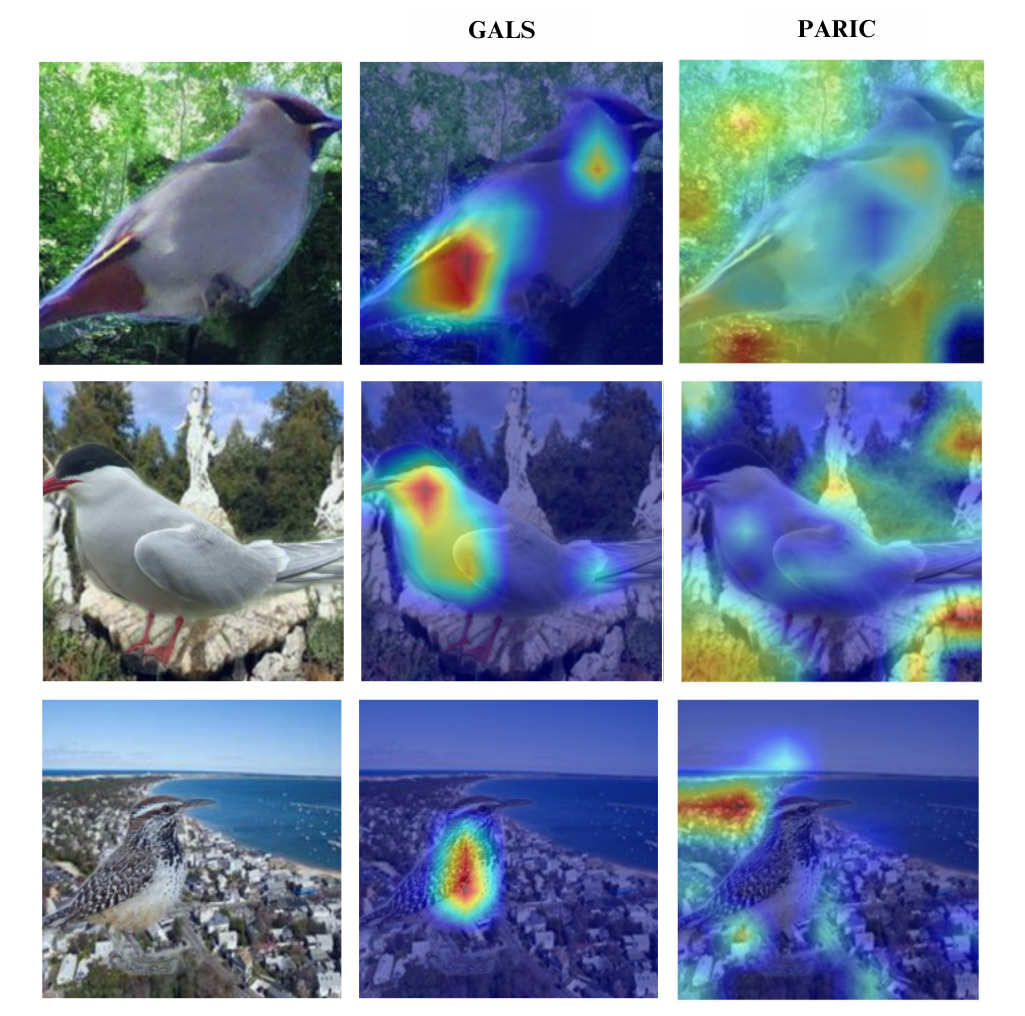}
        \subcaption{Poor attention maps.}
        \label{attmaps3}
    \end{subfigure}
    \caption{\textbf{Comparison of Effective vs. Poor Attention Maps on the Waterbirds 100\% Dataset.} Each subfigure presents three instances from the dataset: the first image is the original input, and the third image is the final attention map obtained after integrating probabilistic layers. In (a), the first two rows use PARIC Mean and the third uses PARIC Median, whereas in (b), the first row uses PARIC Mean and the subsequent rows use PARIC Median.}
    \label{fig:combined_waterbirds}
\end{figure*}

\begin{figure}[htbp]
    \centering
    \includegraphics[width=0.48\columnwidth]{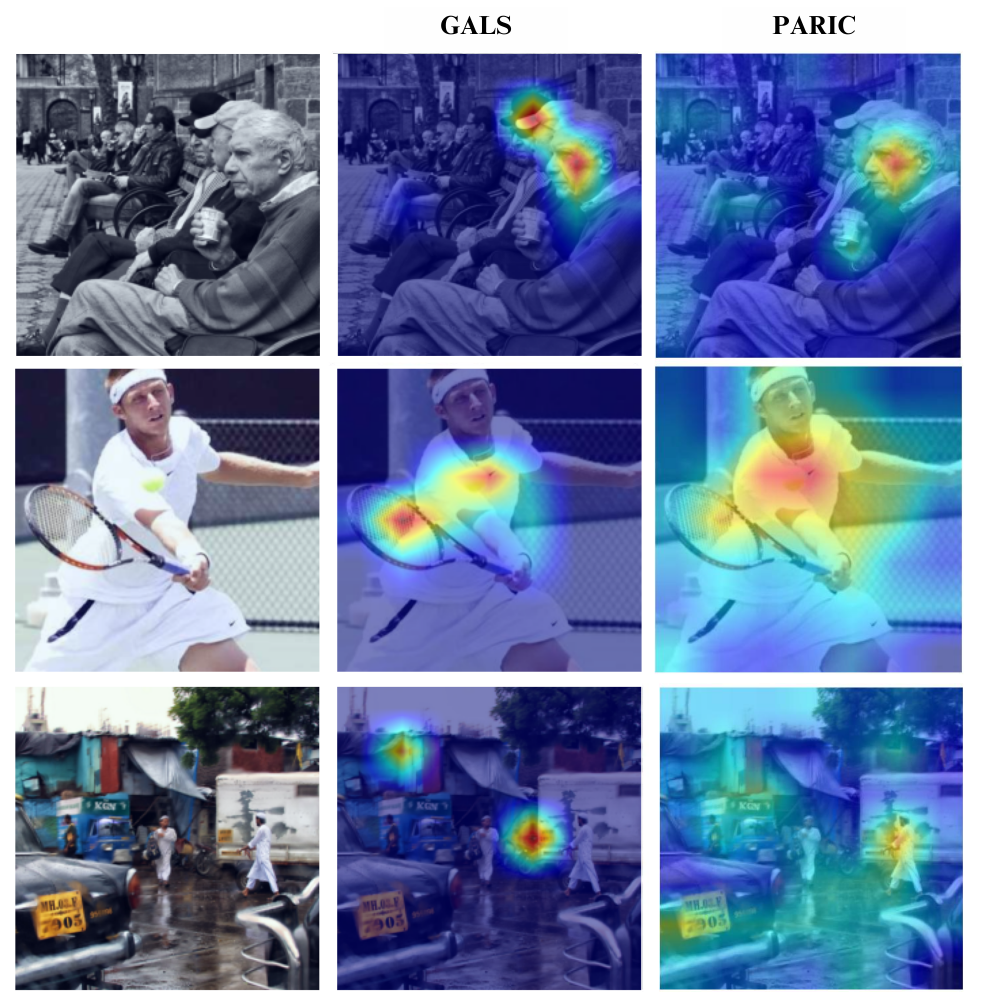}
    \vspace{0.5em}
    \includegraphics[width=0.48\columnwidth]{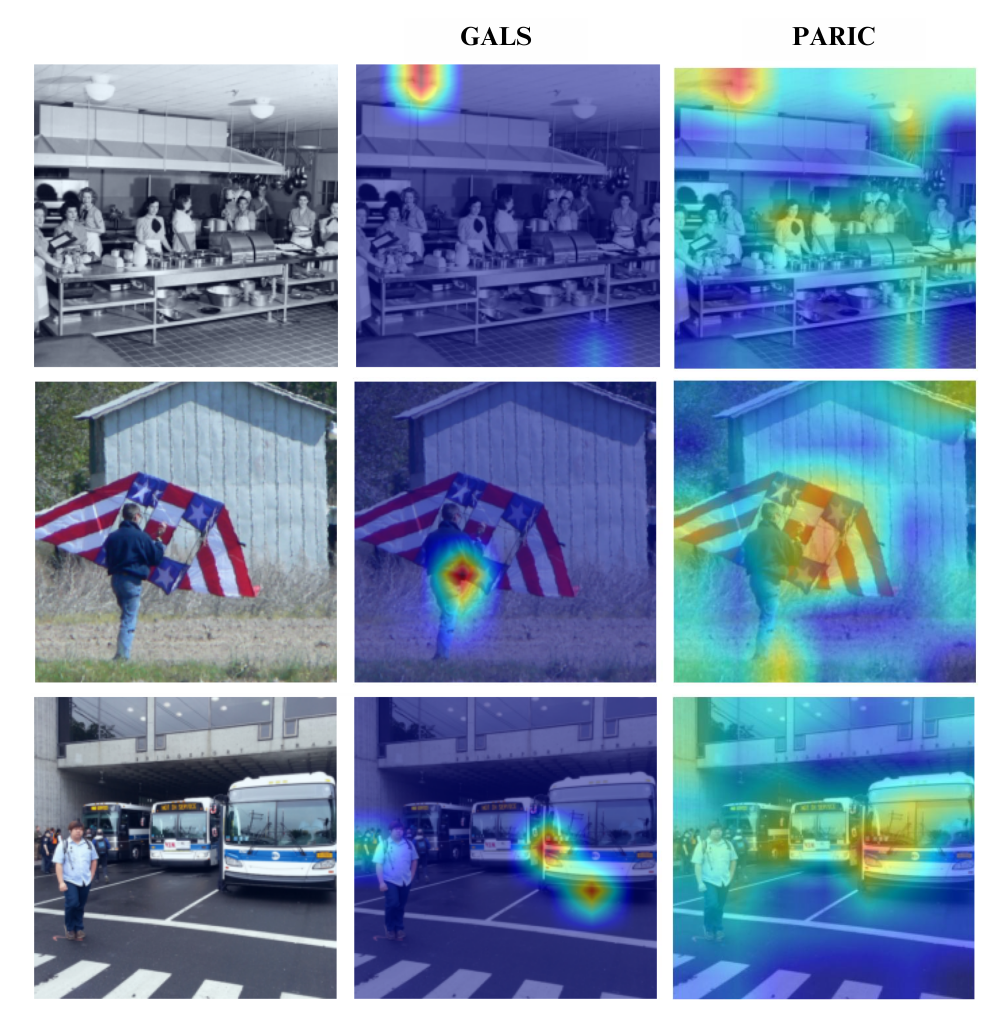}
    \caption{\textbf{\textit{Attention maps for MSCOCO}}. In each row, the first image represents the original input, followed by the attention map from the frozen CLIP model. The third image shows the attention map obtained after integrating probabilistic layers and sampling 50 embedding using the mean aggregation method.}
    \label{fig:attmaps_extra}
\end{figure}

\paragraph{Attention maps visualization}
The attention maps in Figs. \ref{attmaps}, \ref{attmaps2} and \ref{fig:attmaps_extra} visually demonstrate how probabilistic embeddings enhance model interpretability. PARIC's attention maps focus on semantically meaningful regions, reducing variability compared to deterministic approaches. For example, in the Waterbirds dataset, PARIC captures the entire bird even when background cues conflict with class labels. Similarly, in the COCO dataset, PARIC produces broader, more context-aware attention maps that focus on the person rather than disjointed body parts, mitigating overfitting and improving consistency. Despite these advantages, the proposed framework has limitations. In some instances like the example attention map for Food-101 (\autoref{attmaps} and Fig. \ref{attmaps3}, bottom two rows), the regularization can be too strong, limiting model expressiveness. Both guided approaches also rely on CLIP’s deterministic embeddings as a foundation, meaning the performance depends on the quality of CLIP’s representations. If CLIP fails to capture relevant features, GALS and PARIC may also struggle.

%% file: sec/6_discussion.tex
\section{Discussion}
\label{sec:discuss}

PARIC demonstrates improvements owing to the probabilistic approach in specific cases involving imbalanced and noisy datasets, by reducing variability and providing more consistent and equitable predictions. In the Waterbirds dataset, PARIC effectively reduces the influence of background correlations, enabling the model to focus on the bird’s features. This leads to more robust generalization, especially in challenging scenarios where spurious biases may mislead predictions. In the COCO Gender dataset, PARIC mitigates gender biases, offering more balanced results across classes and reducing variability. Leveraging its probabilistic embeddings, PARIC increases stability and fairness, making it a reliable framework for biased or imbalanced datasets.

In PARIC, mean aggregation provides strong overall performance by smoothing predictions across runs, leading to improved generalization. However, it does not account for potential variability in predictions caused by ambiguous or biased data. Median aggregation, on the other hand, is better suited for mitigating extreme variations, resulting in more stable and consistent predictions. This stability is particularly beneficial in datasets like COCO Gender, where fairness metrics are critical for underrepresented classes.

Integrating uncertainty-aware aggregation into PARIC could further harness the strengths of its probabilistic embeddings. Dynamically weighting predictions based on confidence or uncertainty measures can enable PARIC to better adapt to noisy or ambiguous data. For instance, in a dataset prone to inherent subjectivity, such as COCO Gender, incorporating uncertainty measures could help the model identify and prioritize ambiguous cases, further improving fairness and robustness. Additionally, as future work, increasing the number of samples generated from probabilistic encoders could refine the attention maps. More accurate and detailed attention maps would offer better guidance for the classifier, potentially enhancing both interpretability and predictive performance. This approach would allow PARIC to fully leverage its probabilistic nature for improved consistency, fairness, and overall reliability.

%% file: sec/7_conclusion.tex
\section{Conclusions}
\label{sec:conclusions}
This work introduces a natural language-guided attention framework leveraging probabilistic embeddings adapted from pre-trained vision-language models such as CLIP. The probabilistic treatment of embeddings allows treatment of intrinsic multivaluedness and ill-posedness of cross-modal mappings. The proposed probabilistic framework of guiding visual attention through language specifications is validated on three challenging benchmark test problems consisting of noise, implicit bias and class imbalance, where it delivers improved classification performance and model stability in a majority of cases.